\documentclass[preprint,12pt,authoryear]{elsarticle}

\usepackage{amssymb}
\usepackage{amsthm}
\usepackage{natbib}
\usepackage[left=2.5cm, right=2.5cm, top=3cm, bottom=3cm, footskip=0.5cm]{geometry}
\usepackage{graphicx}
\usepackage{hyperref}
\usepackage{booktabs}
\usepackage{amsfonts}
\usepackage{amsmath}

\journal{ }
\begin{document}

\begin{frontmatter}

\title{Towards Ontology-Based Descriptions of Conversations with Qualitatively-Defined Concepts}
\date{ }

\author[1]{Barbara Gendron}
\author[1,2]{Gaël Guibon}
\author[1]{Mathieu d'Aquin}

\address[1]{Loria, CNRS, Université de Lorraine, Nancy, France}
\address[2]{Université Sorbonne Paris Nord, CNRS, Laboratoire d'Informatique de Paris Nord, LIPN, F-93430 Villetaneuse, France}
\address{\textnormal{\texttt{first\_name.last\_name@loria.fr}}}

\end{frontmatter}

\section{Introduction}
\label{sec:intro}

Conversational agents based on Large Language Models (LLMs) have emerged as the preferred approach in contemporary dialogue systems. Their prominence stems from being trained on massive textual corpora, which enables them to generate responses across a wide range of topics with remarkable fluency in user-agent conversations. However, the conversational abilities of LLM-based agents essentially lie in their ability to represent and render knowledge. Their inherent black-box nature prevents the output from being fully predictable, which raises some concerns regarding the generation of inaccurate information or even hallucinations. Moreover, the generated content is not personalized. This is because the model has not been provided with any explicit instructions regarding the specificities of user interaction, whether these are based on personality or task.

In order to enhance user trust in conversational agents without compromising response accuracy, generation control is increasingly considered as a promising solution~\citep{surveycontrol}. In addition to offering perspectives to bring personalization in a predictable fashion, generation control, usually achieved in the framework of constrained generation, has been shown to efficiently improve generation quality~\citep{sun-etal-2023-evaluating}. Since we want more predictable and relevant conversational agents, in our case this control has to be formally defined, consistently applied, and oriented towards human knowledge. Inspired by the advent of neurosymbolic AI and hybridization approaches between LLMs and Knowledge Graphs (KGs) in tasks such as question answering~\citep{fatemi:2023} or planning~\citep{gloria-silva-etal-2024-plan}, we aim here to explore how generation control in specific use-cases can be ontologically defined.

In this work, we focus on enabling conversational control leveraging logical definitions of characteristics of a conversation as established in a dedicated ontology. \citet{varshney:2023} demonstrate that the relevant aspects to consider in order to achieve personalization are user-oriented aspects, and such features are mostly qualitatively defined (e.g., user emotion, conviction, satisfaction, language level, etc.). These features provide key insights into user-agent interactions, but they lack a formal, unique, and commonly approved quantitative definition, which makes it impossible to directly integrate their definition in ontologies in a way that can be used to control conversations.\\

\noindent \textbf{Proposed Approach}. To overcome this issue, we propose an approach based on designing a quantitative definition of qualitatively-defined conversational features by leveraging selected pre-defined descriptors. Once the relevant descriptors are identified, we extract subclass membership rules for each feature based on descriptor values. We then express these rules in description logic so that they can be integrated into an ontology to benefit from consistency checking and inference capabilities. By integrating the ontology information after a reasoning step into the prompts during an LLM fine-tuning step, we manage to control generation on the defined conversation features. \\

\noindent \textbf{Use-Case}. In the following parts, we summarize the approach and results in a specific use-case: \textsl{Proficiency-Level Control}. It consists of modeling the English proficiency of the user and consequently adapting the agent's language level. 

\section{General Framework}

Let's assume that we have, as a conversation aspect, a concept $C$ that has $n$ sub-concepts $\tilde{C} = \left\{ C_i, \enspace i \in \textlbrackdbl  1,n  \textrbrackdbl\right\}$. This concept's definition relies on qualitative, subjective criteria.

We also have a dataset of $N$ texts annotated with the sub-concepts of $C$: 
\begin{equation*}
    D_C = \left \{ (t_i, c_i), \enspace t_i \in T, \enspace c_i \in \tilde{C},\enspace i \in \textlbrackdbl  1,N  \textrbrackdbl\right\}
\end{equation*}

From the annotated dataset \( D_C \), the objective is to derive a quantitative definition of the concept \( C \) based on a set of quantitative descriptors. Let \( \mathcal{F} = \{f_1, f_2, \dots, f_m\}\) represent the selected descriptors for the analysis, where each feature \( f_j \) corresponds to a measurable property of the texts in \( D_C \).

For each sub-concept \( C_i \in \tilde{C} \), we use a classifier to determine from the annotated data a range of values \( R_j^{C_i} \) of each descriptor \( f_j \) which provides a reliable description of $C_i$-annotated texts:
\begin{equation*}
    R_j^{C_i} = [\min(f_j | c = C_i), \max(f_j | c = C_i)], \quad \forall f_j \in \mathcal{F}.
\end{equation*}

The quantitative definition of \( C_i \) is then represented as the conjunction of conditions that a text must satisfy for all descriptors:
\begin{equation*}
Q(C_i) = \bigwedge_{j \in \{1, \dots, m\}} \left( R_j^{C_i} \right) \quad \text{where} \quad R_j^{C_i} = [\min(f_j | c = C_i), \max(f_j | c = C_i)]
\end{equation*}

To evaluate a new text \( t \in T \) with respect to the concept \( C \), the values of the descriptors \( \mathcal{F} \) are computed, and the text is assigned to the sub-concept \( C_i \) whose quantitative definition \( Q(C_i) \) matches the extracted values.

\section{Proficiency Level Modeling}
\label{sec:method-plm}

Here, we explore how the framework described above can be used and tested on language proficiency levels with the aim of ensuring that a conversation remains at a level comfortable to the user.

To model proficiency levels, we rely on the Common European Framework of Reference for Languages\footnote{\url{https://www.coe.int/en/web/common-european-framework-reference-languages/table-1-cefr-3.3-common-reference-levels-global-scale}} (CEFR). There are six CEFR levels divided into three categories~\citep{North2007}: A1 and A2 are referred to as ``basic user'' levels, B1 and B2 as ``independent user'' levels, and C1 and C2 as ``proficient user'' levels. Each level is defined using qualitative statements, often subjective in nature, such as ``can understand the main ideas of complex texts in both concrete and abstract topics'' (B2), or ``can produce simple connected texts'' (B1). Besides, CEFR levels are originally expressed in terms of the user's proficiency, from which we can naturally derive a definition for an utterance's proficiency level. For instance, a B1-level utterance is an utterance that can be phrased by a user whose proficiency level is exactly B1.

Therefore, the proficiency-level control task requires working on concepts that derive from other qualitatively-defined concepts. This is what makes proficiency level assessment so challenging, as highlighted by~\cite{arase-etal-2022-cefr} where they provide a high-quality dataset of CEFR-annotated sentences for which we calculate an inter-annotator agreement score of 0.41 based on the accuracy between the given labels, in order to obtain a comparative topline.  

In that sense, providing consistent, factually-defined (quantitative) definitions of CEFR levels would not only allow for these definitions to be used in an ontological framework, but is also expected to help overcome discrepancies in annotations.

Regarding feature choice, the approach is to consider any linguistic feature that can describe the complexity of the text or a related aspect. It is first natural to consider readability metrics since they are specifically designed for that. For instance, we use Gunning-Fog \citep{gunning1969fog}, Flesch-Kincaid~\citep{flesch1948new} and Dale-Chall~\citep{dale1949concept} readability scores. In addition to direct measures of text complexity, we also take into account linguistic features. For the sake of completeness, we consider features describing a wide range of linguistic properties: lexical (e.g. named entity count, average word length), morphological and syntactic (e.g. coordination, conjunctions, subordinations) properties are included. We also consider discourse features related to rhetoric and connectivity (e.g. pronoun density, presence of indirect speech). All these features are either binary or expressed as a numerical score.

\section{Experiments}

In this part, we show how our general framework applies to the Proficiency-Level Control use-case. In this context, our framework comprises the following input variables: 
\[
C=\texttt{ProficiencyLevel} \qquad \tilde{C} = \left\{ \text{A1}, \text{A2}, \text{B1}, \text{B2}, \text{C1}, \text{C2}\right\}.
\]

\noindent\textbf{Model.} In order to establish the relation between the numerical values of descriptors and the qualitative definition of proficiency levels, we choose to train a Decision Tree Classifier (DTC) \citep{quinlan1986induction,breiman2017classification} model to perform CEFR-level prediction. An advantage of such a model is that it provides explicit belonging rules for each class based on the selected features. Another advantage is that feature selection can be further deduced after the training phase by computing feature importance and implementing a simple threshold strategy. Using Gini impurity~\citep{Breiman1984} as a selection criterion, we restrict class definitions to a limited number of features (descriptors) with minimal impact on performance. Moreover, we limit the tree depth to 5 and require a minimum of 50 samples to form a new branch. These constraints help to mitigate overfitting, which would lead to non-generalizable definitions.\\

\noindent\textbf{Data.} To train the classifier and derive high-quality definitions, we require diverse CEFR-annotated textual data. For this, we selected two datasets: CEFR-T, extracted from~\cite{nallapati-etal-2016-abstractive} CNN/DailyMail dataset, which contains expert-annotated news texts, and CEFR-SP~\citep{arase-etal-2022-cefr}, which consists of CEFR-labeled single sentences. To determine which dataset yields the most relevant definitions, we experimented with training on one and testing on the other. The results showed that training on CEFR-T led to the best performances on test. Additionally, training on both datasets produced overly specific rules that failed to capture the distinctions between the two data types. Consequently, we opted to train the classifier exclusively on CEFR-T while using CEFR-SP as a control dataset.\\

\begin{figure}
    \centering
    \includegraphics[width=\linewidth]{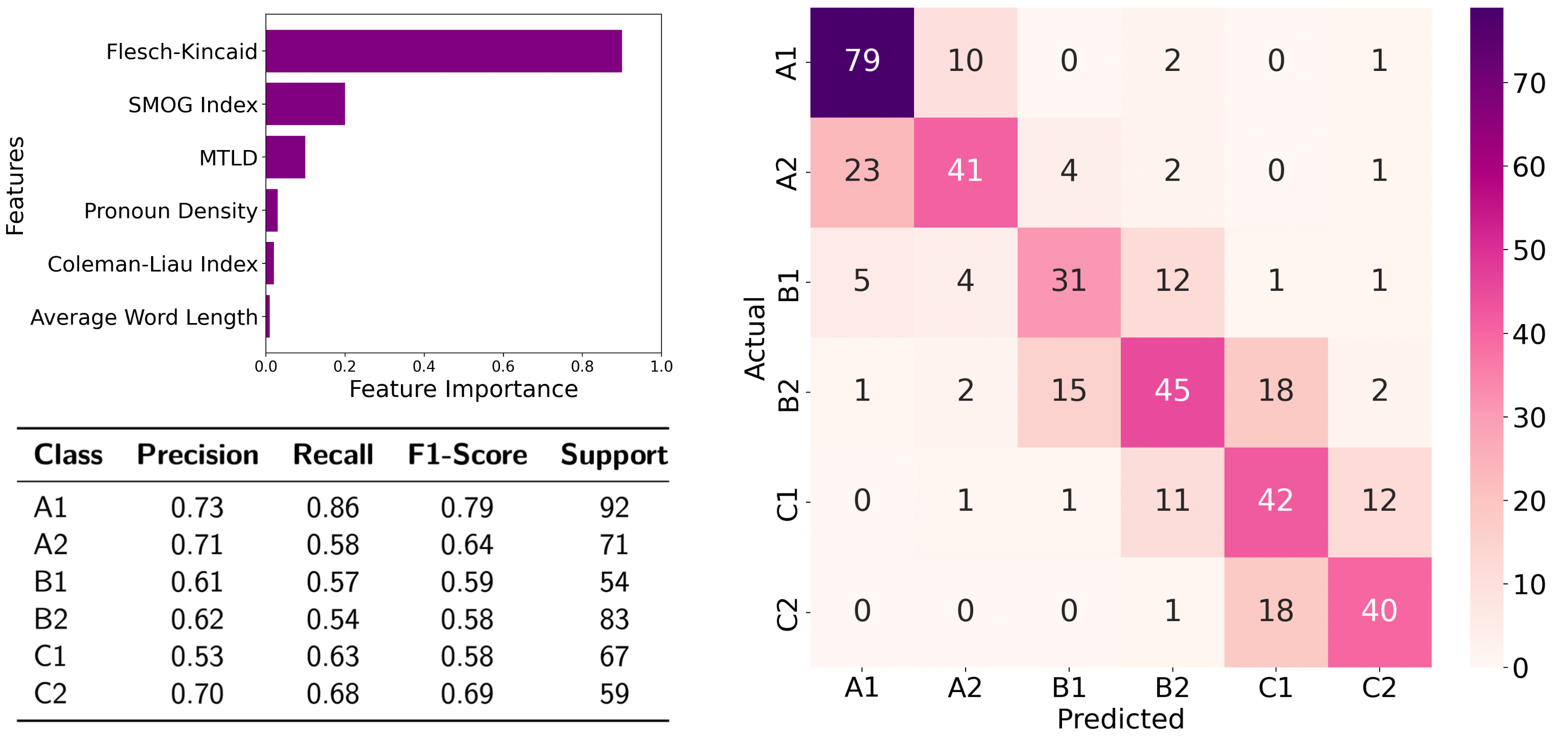}
    \caption{\centering Feature importances \textsl{(top left)}, classification metrics \textsl{(bottom left)} on CEFR-T validation set. On the right, the confusion matrix.}
    \label{fig:dt_res}
\end{figure}

\noindent\textbf{Results.} All the insights about the classification results are gathered in Figure~\ref{fig:dt_res}. Our model achieves an overall accuracy of 0.66, which is promising considering the inter-annotator agreement score of 0.41 mentioned in Section~\ref{sec:method-plm}. 
In this specific use case, since the sub-concepts are ordered, it makes sense to evaluate the relevance of the DTC by computing the Mean Absolute Error (MAE) score: we obtain a value of 0.42. This is consistent with the confusion matrix in Figure~\ref{fig:dt_res}, where it is visible that most of the errors consist of misclassifying between two neighboring classes.

From the DTC, we explore the tree to identify the feature value ranges for each class, treat redundancies, and translate these range constraints to a description logic syntax -- namely the Manchester syntax \citep{horridge2006manchester} --. The proficiency level sub-concepts are then defined in an ontology using Protégé~\citep{protege}. Concretely, we define a concept \texttt{Utterance} with sub-concepts \texttt{A1LevelUtterance} to \texttt{C2LevelUtterance}.\\

\section{Related Work}
\label{sec:sota}

State-of-the-art approaches for CEFR-level classification on textual data leverage deep learning architectures like Recurrent Neural Networks~\citep{bert-cefr-1} or based on BERT ~\citep{devlin-etal-2019-bert} model~\citep{bert-cefr-2}. Although these models achieve respectively 0.75 and 0.95 classification accuracy on the EFCAMDAT dataset~\citep{efcamdat}, their approaches are not comparable to ours for two main reasons. First is the dataset, as the major part of existing work on CEFR-level classification relies on the EFCAMDAT. However, as highlighted in~\cite{ballier:hal-02496670}, EFCAMDAT labels are not directly CEFR levels, and provides a significantly imbalanced dataset when its labels are mapped to CEFR levels. Second is the black-box nature of deep learning models, which does not allow to end up with formal rules of belonging to each class. Therefore, such approaches are not compatible with our use-case.

A more similar approach has been proposed by~\cite{Gaillat_Simpkin_Ballier_Stearns_Sousa_Bouyé_Zarrouk_2022} and consists in employing machine learning techniques to represent each class through a set of well-chosen linguistic features. This approach also leverages the EFCAMDAT dataset and achieves 0.82 classification accuracy. However, it appears that the model performs very differently depending on the class, with class-wise F1 scores ranging from 0.03 to 0.90. Hence, it appears that the approach struggles to produce definitions of an equivalent quality across proficiency levels.

\section{Perspectives}

\begin{figure}
    \centering
    \includegraphics[width=\textwidth]{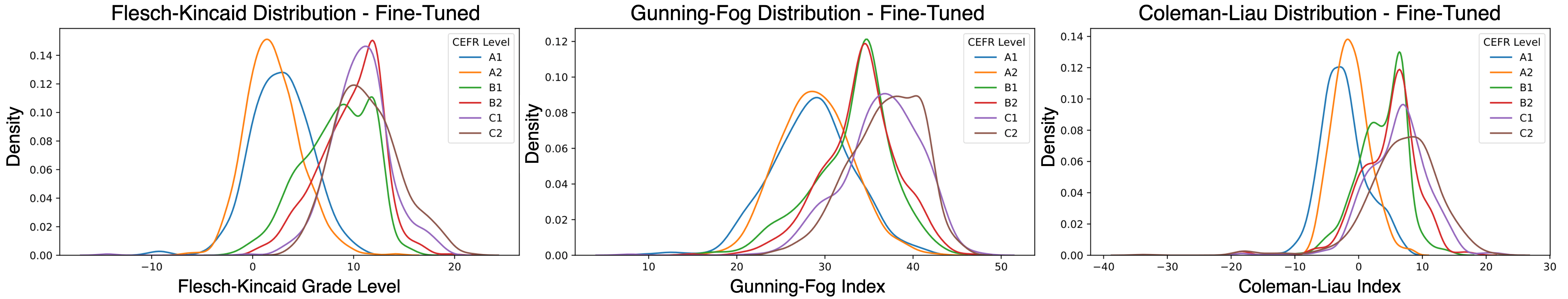}
    \caption{\centering Readability Metric Distributions per Predicted CEFR Level}
    \label{fig:feat}
\end{figure}

As mentioned in the Introduction, the motivation of our approach is to perform ontological control on LLM-generated outputs. This can be done by using the DTC-inferred ontological definitions of CEFR levels to annotate other datasets, on which language models can be trained. Therefore, the next part of our work is to implement a fine-tuning procedure, relying on Causal Language Modeling, during which the language model progressively learns how to generate according to the asked ontological concept. We applied this procedure on \texttt{Llama3-8B-Instruct}~\citep{touvron2023llamaopenefficientfoundation} model along with LoRA adapters~\citep{lora-adapters} for parameter-efficient fine-tuning~\citep{houlsby}. Figure~\ref{fig:feat} shows how the generated content differentiates along the readability metrics used in ontological definitions. The top line uses the raw model, without fine-tuning, while the bottom line uses our fine-tuned model. It shows that our fine-tuning procedure provides a better distinction of the Flesch-Kincaid index (FKGL) between CEFR levels (FKGL has the highest feature importance in our decision tree classifier). With the pre-trained model, the FKGL values have similar distributions across the CEFR levels, whereas with the fine-tuned model, the FKGL discriminates simple from medium and complex sentences.

\begin{figure}
    \centering
    \includegraphics[width=\textwidth]{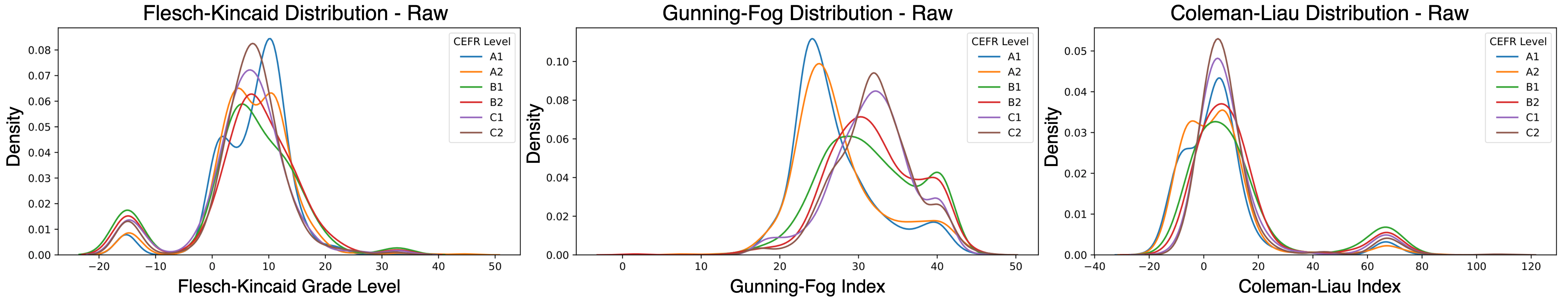}
\end{figure}

\section{Conclusion}

This work introduces an ontology-based approach for controlling conversational generation by quantitatively defining qualitatively-described conversation aspects. By leveraging linguistic descriptors in a decision tree classifier, we establish structured definitions that can be integrated into an ontological framework, ensuring consistency and transparency. The use-case of Proficiency-Level Control demonstrates the effectiveness of this approach in modeling CEFR levels, which are originally qualitatively-defined concepts. Experimental results indicate that the proposed method provides relevant definitions of concepts despite the inherent subjectivity in the annotation of such concepts.

Future directions include refining the ontological definitions, extending the methodology to additional qualitative aspects of conversation, and refining the fine-tuning procedure to improve its efficiency and eventually to enable control on several aspects at the same time by combining definitions. Eventually, we aim to enhance the reliability and adaptability of conversational large language models, making interactions more transparent and tailored to user needs.

\bibliographystyle{elsarticle-num-names} 
\bibliography{cas-refs}

\end{document}